\documentclass{article}

\usepackage{fullpage,times,color,hyperref}

\usepackage{amsmath,amsthm,amssymb}
\usepackage{natbib}
\usepackage{mathtools}
\newtheorem{theorem}{Theorem}

\newtheorem{example}[theorem]{Example}

\newenvironment{definition}[1][Definition]{\begin{trivlist}
\item[\hskip \labelsep {\bfseries #1}]}{\end{trivlist}}

\title{Elimination of Spurious Ambiguity \\
  	in Transition-Based Dependency Parsing}
\author{
  Shay B. Cohen\\
  Department of Computer Science \\
  Columbia University, USA \\
  \texttt{scohen@cs.columbia.edu} \and
  Carlos G{\'o}mez-Rodr{\'\i}guez\\
  Departamento de Computaci\'{o}n \\
  Universidade da Coru\~{n}a, Spain \\
  \texttt{cgomezr@udc.es} \and
  Giorgio Satta\\
  Department of Information Engineering \\
  University of Padua, Italy \\
  \texttt{satta@dei.unipd.it}}

\date{}

\mathtoolsset{showonlyrefs,showmanualtags}

\newcommand{\eat}[1]{}
\newcommand{\termdef}[1]{\textbf{#1}}
\newcommand{\size}[1]{\left| {#1} \right|}
\newcommand{\sep}{\,\mid\,}

\newcommand{\mytrue}{{\textrm{\texttt{T}}}}
\newcommand{\myfalse}{{\textrm{\texttt{F}}}}

\renewcommand{\labelenumi}{\rm{(\roman{enumi})}}

\newcommand{\init}{I}

\newcommand{\myalph}{\mathit{\Sigma}}

\newcommand{\stack}{\sigma}
\newcommand{\buffer}{\beta}
\newcommand{\config}[3]{{( {#1} , {#2} , {#3} )}}

\newcommand{\transname}[1]{\ensuremath{\mathsf{#1}}}

\newcommand*{\map}[2]{#1\mathpunct{:} #2}
\newcommand*{\pf}{\rightharpoonup}

\newcommand{\fstop}{\mathsf{stop}}

\newcommand{\fredl}[1]{\mathsf{left}_{#1}}
\newcommand{\fredr}[1]{\mathsf{right}_{#1}}

\newcommand{\pgenredl}[4]{\mathsf{left}({#1},{#2};{#3},{#4})}
\newcommand{\pgenredr}[4]{\mathsf{right}({#1},{#2};{#3},{#4})}

\newcommand{\depth}[1]{\mathrm{depth}({#1})}
\newcommand{\degree}[1]{\mathrm{deg}({#1})}

\newcommand{\tshift}{sh}

\newcommand{\treduceleft}{la}
\newcommand{\treduceright}{ra}

\newcommand{\tshiftleaf}{sh}
\newcommand{\tshiftnonleaf}{sh}
\newcommand{\treduceleftleaf}{la}
\newcommand{\treduceleftnonleaf}{la}
\newcommand{\treducerightleaf}{ra}
\newcommand{\treducerightnonleaf}{ra}

\newcommand{\pbu}[2]{\ensuremath{\mathsf{bu}({#1},{#2})}}

\newcommand{\reduceleft}[2]{\ensuremath{\transname{\treduceleft}_{{#1} \leftarrow {#2}}}}
\newcommand{\reduceright}[2]{\ensuremath{\transname{\treduceright}_{{#1} \rightarrow {#2}}}}

\newcommand{\shift}{\ensuremath{\transname{\tshift}}}

\newcommand{\reduceleftleaf}[2]{\ensuremath{\transname{\treduceleftleaf}^{s}_{{#1} \leftarrow {#2}}}}
\newcommand{\reducerightleaf}[2]{\ensuremath{\transname{\treducerightleaf}^{s}_{{#1} \rightarrow {#2}}}}
\newcommand{\reduceleftnonleaf}[2]{\ensuremath{\transname{\treduceleftnonleaf}^{\overline{s}}_{{#1} \leftarrow {#2}}}}
\newcommand{\reducerightnonleaf}[2]{\ensuremath{\transname{\treducerightnonleaf}^{\overline{s}}_{{#1} \rightarrow {#2}}}}

\newcommand{\shiftnonleaf}{\ensuremath{\transname{\tshiftnonleaf}^{\overline{s}}}}
\newcommand{\shiftleaf}{\ensuremath{\transname{\tshiftleaf}^{s}}}

\newcommand{\conftransshort}[3]{#1 : #2 \vdash #3 \; }

\newcommand{\conftrans}[3]{#1 : & #2 \vdash  \\ & #3 \; }

\newcommand{\compbytrans}[2]{{\ensuremath{(#1;#2)}}}

\newcommand{\mfunction}[1]{\mu({#1})}
\newcommand{\myhom}{\tau}
\newcommand{\transftransition}[2]{\myhom_{#1}({#2})}

\newcommand{\INFERRULE}[3][]{%
	\ensuremath{\inferrule{\mathstrut #2}{\mathstrut #3}\ifx\\#1\\\else\enskip(#1)\fi}}

\newcommand{\treeof}[1]{\ensuremath{D({#1})}}

\newcommand{\ignore}[1]{}

\newcommand{\order}[1]{{\cal O}({#1})}

\def\newcite{\cite}
\def\namecite{\newcite}

\date{}

\begin{document}
\maketitle

\begin{abstract}

We present a novel technique to remove spurious ambiguity from transition systems for dependency parsing.
Our technique chooses a canonical sequence of transition operations (computation) for a given dependency tree.
Our technique can be applied to a large class of bottom-up transition systems, including for instance \namecite{nivre2004incrementality} and \namecite{attardi2006experiments}.
\end{abstract}


\section{Introduction}
\label{sec:intro}

In parsing, spurious ambiguity refers to ambiguity in a grammar that occurs because several derivations exist for an identical syntactic analysis. 
When the grammar is enriched with probabilities, the existence of spurious ambiguity implies that the statistical model is defined over {\em derivations}, a more fine-grained version of the actual syntactic structures of interest. The probability of a syntactic structure then becomes the marginalized probability over all derivations that
map to that syntactic structure.

Spurious ambiguity can exist in various grammatical models such as 
combinatory categorial grammars 
\citep{steedman01:_syntac_proces},
tree adjoining grammars \citep{joshi1975tree},
data-oriented parsing \citep{bod-92} and transition-based dependency parsing \citep{nivre2005dependency}.

While models with spurious ambiguity are statistically more expressive than models without spurious ambiguity,\footnote{By this we mean that 
there are distributions over syntactic structures which can be obtained using models with spurious ambiguity but can not be obtained using models without spurious ambiguity.} an obstacle exists in the need to marginalize out derivations in order to compute the total probability of a syntactic structure, which is necessary for training and decoding with such models. For many models with spurious ambiguity, it is in fact provably NP-hard to do such marginalization \citep{simaan-96}. 

Various heuristics exist to sidestep the need for marginalization. For example, during decoding, one can find the highest-scoring derivation instead of the highest-scoring structure. Under the assumption that most of the probability mass of a given syntactic structure is concentrated on a single derivation, this alternative decoding can be successful.  However, this assumption often fails when the probability mass is evenly divided for one syntactic structure but concentrated on a single derivation for another. Even when marginalization can be done efficiently, the likelihood of observed data often becomes non-convex, which is undesirable for training the model because of the local optima problem. For these reasons, it is preferable in most cases to eliminate spurious ambiguity.

In this paper, we focus on eliminating spurious ambiguity that exists in transition-based dependency parsing. Ambiguity arises because several sequences of shift and reduce operations (which assemble a derivation) could yield identical dependency trees. The transition-based parsing literature has implicitly tackled the issue of spurious ambiguity by defining an {\bf oracle} which, after receiving a dependency tree as input, outputs a unique derivation for that tree based on a canonical ordering of the transition operations. This oracle is then used on the training data (pairs of sentences and dependency trees), yielding new training data (pairs of sentences and shift-reduce derivations) to train multi-class classifiers that decide at each transition step which operation to take \citep{nivre04memorybased}.

Rather than eliminating spurious ambiguity from the model, this heuristic creates a bias through training to prefer certain derivations for a given dependency tree when doing decoding.
In addition, as we discuss in \S\ref{sec:experiment}, some of the existing oracles for supervised dependency parsing are based on incomplete heuristics (which are often undocumented).

We present a more principled approach to eliminate spurious ambiguity in transition-based dependency parsing. We first define a wide class of bottom-up transition systems, which includes the arc-standard transition system \citep{nivre2008algorithms} as well as the transition system from \newcite{attardi2006experiments}. 
One could also define a transition-based parser using a strategy which is a hybrid between the arc-standard strategy and the easy-first strategy from \newcite{goldberg-elhadad:2010:NAACLHLT}, in which a set of shift actions would need to be taken before a reduction decision is made affecting elements at some deeper position on the stack: this decision can depend on the ``easiness'' of the reduction.  Such a parser can be easily encapsulated into our framework. 

We then provide a general technique to enrich the transitions of these systems in order 
to remove spurious ambiguity while maintaining the completeness of the enriched 
system with respect to the original. Each tree is associated with a single derivation, 
which is a sequence of shift and reduce operations such  that reduce operations are performed as soon as possible, and conflicts between several reductions are resolved by first attaching dependents that are closer to the current focus point of the parser (top of the stack). This is coherent with psycholinguistic models postulating that humans tend to process local attachments first \citep{Gibson2000}.

Our approach eliminates ambiguity from a declarative transition system. However, it is extensible to a decoding algorithm as well. The transition systems we introduce can be made
probabilistic in a manner similar to the one that appears in \newcite{cohen2011exact}. Then, a dynamic programming algorithm for these probabilistic systems can be derived so that one can identify the highest scoring derivation and compute the expectations of features in the model \citep{kuhlmann2011acl,cohen2011exact}. Our removal of spurious ambiguity is efficient: the
dynamic programming algorithm which is based on the transformed transition system has the same asymptotic complexity as a dynamic programming algorithm for the original transition system.

Our original motivation was to construct a probabilistic model for transition-based dependency parsing, such that a unique (canonical) derivation exists for each dependency tree.  This avoids the computational complexity involved in marginalizing derivations.   Removal of spurious ambiguity in such a case has to be done at the level of the transition system and not at the level of a tabular method simulating the system or at the level of the resulting parse forest: removing undesired derivations from the chart does not tell us how to set transition probabilities in the original system in such a way that the probability mass of each dependency tree is allocated to a single canonical derivation.

The rest of this paper is organized as follows. We provide an overview of transition-based dependency parsing in \S\ref{sec:def}. We then
describe the main details of the spurious ambiguity removal technique in \S\ref{sec:removal}. We provide proofs and formal analysis in \S\ref{sec:formal}.
We apply our technique to the parser from \newcite{attardi2006experiments} and run 
some experiments in \S\ref{sec:experiment}.
We describe other applications of our technique in \S\ref{sec:discussion},
and we conclude with an open problem in \S\ref{sec:conclusion}.

\section{Transition-Based Dependency Parsing}
\label{sec:def}

In this section we briefly introduce the basic definitions for transition-based dependency parsing; we refer the reader to \cite{nivre2008algorithms} for a more detailed presentation.  We also define the class of transition-based parsers which is investigated in this paper.

\subsection{General Transition Systems}
\label{ssec:trans}

Let $\myalph$ be an input alphabet and let 
$w = a_1 \cdots a_{n}$, $n \geq 1$, be the input string 
with $a_i \in \myalph$ for each $i$ with $1 \leq i \leq n$.
A \termdef{dependency tree} for $w$ is a directed 
tree $G = (V_w, A)$ where $V_w = \{0, 1, \dots, n\}$ is the 
set of nodes and $A \subseteq V_w \times V_w$ is a set of 
arcs. Each node encodes the position of a token 
in~$w$, with $0$ being a dummy node used as an artificial root, and each arc
encodes a dependency relation between two tokens.  We write $i \to j$ to denote 
a directed arc $(i, j) \in A$, where node~$i$ is the 
head and node~$j$ is the dependent.

A \termdef{transition system} for dependency parsing is a tuple $S = (C, T,
\init, C_t)$, where~$C$ is a set of configurations, defined below, $T$~is a
finite set of \termdef{transitions}, which are partial 
functions $\map{t}{C \pf C}$, $\init$~is a total initialization 
function mapping each input string to a unique initial configuration,
and $C_t \subseteq C$ is a set of terminal configurations.

A \termdef{configuration} is defined relative to input string $w$, 
and is a triple $\config{\stack}{\buffer}{A}$.  Symbols $\stack$ 
and~$\buffer$ are disjoint lists of nodes from $V_w$, 
called \termdef{stack} and input \termdef{buffer}, respectively, 
and $A \subseteq V_w \times V_w$ is a set of arcs.  If $t$ 
is a transition and $c_1, c_2$ are configurations such 
that $t(c_1) = c_2$, we write $c_1 \vdash_t c_2$, or 
simply $c_1 \vdash c_2$ if $t$ is understood from the context.

We denote the stack with its topmost element
to the right and the buffer with its first element to the left.
We indicate concatenation in the stack and buffer by a vertical bar.
For example, for $i \in V_w$, $\sigma|i$ denotes some stack with topmost element
$i$ and $i|\beta$ denotes some buffer with first element $i$.
For $1 \leq i \leq n$, $\beta_i$ denotes the buffer $[i, i+1, \ldots, n]$;
for $i > n$, $\beta_i$ denotes the empty buffer $[]$.

A \termdef{computation} of~$S$ is a sequence $\gamma = c_0, \dots, c_m$,
$m \geq 1$, of configurations such that, for every $i$ with $1 \leq i \leq m$,
$c_{i-1} \vdash_{t_i} c_i$ for some $t_i \in T$.
In other words, each configuration in a computation is obtained
as the value of the preceding configuration under some transition.
A computation can be uniquely specified by its initial
configuration~$c_0$ and the sequence $t_1, \ldots, t_m$ of its transitions.  
Thus we will later denote $\gamma$ in the form 
$\compbytrans{c_0}{t_1, \ldots, t_m}$.

\subsection{Spurious Ambiguity}

A computation $\gamma = c_0, \dots, c_m$ is called \termdef{complete} whenever $c_0 = \init(w)$ for some input string~$w$, and $c_m \in C_t$.  
For a complete computation $\gamma$ we denote as $\treeof{\gamma}$ 
the unique dependency tree 
consisting of nodes $V_w$ and all arcs in the final configuration $c_m$. 
We say that a transition system has \termdef{spurious ambiguity} if, for some pair of complete computations $\gamma$ and $\gamma'$ with $\gamma \neq \gamma'$, we have $\treeof{\gamma} = \treeof{\gamma'}$.

Informally, the existence of spurious ambiguity implies that there are at least two computations that derive the same dependency tree. Spurious ambiguity exists in various transition systems, such as those in \namecite{nivre2004incrementality} and \namecite{attardi2006experiments}.

\begin{example}
\label{ex:arcstd}
The well-known arc-standard transition system by
\namecite{nivre2004incrementality} can be defined as follows: its initialization
function is $\init(a_1 \cdots a_n) = ([0],[1 \cdots n],\emptyset)$, its set of
terminal configurations is $C_t = ([0],[],A)$, and it has the following
transitions:
\begin{align}
\conftransshort{\transname{shift}}{(\sigma,
i|\beta,A)}{(\sigma|i,\beta,A)} \\
\conftransshort{\transname{la}}{(\sigma|i|j,\beta,A)}{(\sigma|j,\beta,A \cup \{j
\rightarrow i\})} \\ 
\conftransshort{\transname{ra}}{(\sigma|i|j|,\beta,A)}{(\sigma|i,\beta,A \cup
\{i \rightarrow j\})}
\end{align}

The two following complete computations for a string $w = a_1 a_2 a_3$ produce the same tree with arcs $\{ {0 \to 2}, {2 \to 1}, {2 \to 3} \}$:
\begin{enumerate}
  \item[(i)]
  $\compbytrans{\init(w)}{\transname{shift},\transname{shift},\transname{la},\transname{shift},\transname{ra},\transname{ra}}$;
  \item[(ii)]
  $\compbytrans{\init(w)}{\transname{shift},\transname{shift},\transname{shift},\transname{ra},\transname{la},\transname{ra}}$.
\end{enumerate}
Therefore, this transition system has spurious ambiguity, caused by the fact
that it allows words (in the example, $a_2$) to choose whether to collect a left
or a right dependent first.
\end{example}

We remark that while in the case of the arc-standard model spurious ambiguity is restricted to 
a certain set of {\em permutations} over sequences
of operations, i.e., all derivations of a given syntactic tree consist of the same transitions in some permutation, 
this does not hold in the case of non-projective models.

\subsection{Bottom-Up Shift-Reduce Transition Systems}

Many of the transition systems for dependency parsing that 
have been proposed in the literature adopt a bottom-up 
strategy, meaning that they construct dependency trees starting from the leaves
and finishing with the root, by always collecting all the dependents of a
given node before assigning it as a dependent of another node.  
This includes for instance the already mentioned 
arc-standard parser, and
the non-projective parser of \namecite{attardi2006experiments}.
These parsers tend to present spurious ambiguity because, as in Example
\ref{ex:arcstd}, the left and right dependents of a given node can be 
collected in different orders. 
This is in contrast with parsers derived from the arc-eager
model \citep{nivre2003efficient} which are not bottom-up and instead impose
a unique left-to-right order in which arcs must be constructed.

Some bottom-up transition systems
use reduce transitions that affect the buffer, but they can 
be cast in an alternative form in which all reductions involve 
only elements from the stack.  This is done by considering 
the first element of the buffer as the topmost stack symbol, 
as discussed by \namecite{cohen2011exact}; in this way reductions might take 
place between stack elements placed at positions deeper than 
the topmost one. The following definition captures the 
general form of such models. 

\begin{definition}
A transition system is \termdef{bottom-up shift-reduce} if its initialization
function is $\init(a_1 \cdots a_n) = ([0],[1 \cdots n],\emptyset)$, its set of
terminal configurations is $C_t = ([0],[],A)$, and its set of
transitions consists of the following:
\begin{enumerate}
\item[(i)]
a shift transition \transname{\tshift} of the form $(\sigma, i|\beta, A) \vdash (\sigma|i, \beta, A)$; 
\item[(ii)]
a set of left arc transitions \reduceleft{p}{q} with $p > q \geq 1$, each of the form 
\[
(\sigma|i_p|i_{p-1}|\cdots|i_1 , \beta, A) \vdash (\sigma|i_{p-1}|\cdots|i_1, \beta, A \cup \{i_q \rightarrow i_p\});
\]
\item[(iii)]
a set of right arc transitions \reduceright{p}{q} with $p > q \geq 1$, each of the form
\[
(\sigma|i_p|i_{p-1}|\cdots|i_1 , \beta, A) \vdash (\sigma|i_{p}|\cdots|i_{q+1}|i_{q-1}|\cdots|i_1, \beta, A \cup \{i_p \rightarrow i_q\}).
\]
\end{enumerate}
\end{definition} 
Transitions in~(ii) and~(iii) above are called \termdef{reductions}. 
The \termdef{degree} of reductions \reduceleft{p}{q} and \reduceright{p}{q} is defined as
$p-q$ and is always positive. The \termdef{depth} of reductions
\reduceleft{p}{q} and \reduceright{p}{q} corresponds to the index $p$. The
degree of a transition system $S$, written $\degree{S}$, is the maximum degree among all its reductions.
Analogously, the depth of a transition system $S$, written $\depth{S}$, is the
maximum depth among all its reductions.

The next definition introduces a condition that allows us to remove spurious ambiguity from bottom-up shift-reduce parsers.  Informally, the condition requires that the existence in the system of a reduction of some type involving stack positions $p$ and $q$, $p>q$, always implies the existence in the system of reductions of the same type involving stack positions $p'$ and $q'$ with $p'<p$ and $q' \leq q$. 
We need some additional notation.  Let $\mfunction{\reduceleft{p}{q}}$ be a set of transitions including $\reduceleft{p-1}{q}$ if $p>q+1$, $\reduceleft{p-1}{q-1}$ if $q>1$, and no other transition.  Similarly, $\mfunction{\reduceright{p}{q}}$ includes $\reduceright{p-1}{q}$ if $p>q+1$, $\reduceright{p-1}{q-1}$ if $q>1$, and no other transition. 

\begin{definition}
Let $S$ be a bottom-up shift-reduce transition system with set of transitions $T$. $S$ is \termdef{monotonic} if for each $t \in T$ we have $\mfunction{t} \subseteq T$.
\end{definition}

\begin{example}
\label{ex:attardi}
The transition-based parser of \namecite{attardi2006experiments} can be written
as the bottom-up shift-reduce system with transitions
\transname{\tshift}, \reduceleft{p}{1} and \reduceright{p}{1} 
for every $p$ with $2 \leq p \leq d$, $d = \depth{S}$.
The system with 
depth $3$, as used by \cite{Kuhlmann10,cohen2011exact}, has
transitions
\transname{\tshift}, \reduceleft{2}{1}, \reduceright{2}{1}, \reduceleft{3}{1}
and \reduceright{3}{1}.

These systems are monotonic for every value of $d$, 
since for a transition
\reduceleft{p}{1}, we have that $\mfunction{\reduceleft{p}{1}} = \{ \reduceleft{p-1}{1} \}$ 
(if $p>2$) or $\emptyset$ (otherwise), and therefore $\mfunction{\reduceleft{p}{1}}$ is 
included in $T$.  The same also holds for $\mfunction{\reduceright{p}{1}}$.
\end{example}

The monotonicity property is crucial for the main result of this paper: if a bottom-up
shift-reduce transition system is monotonic, we can systematically obtain an equivalent system
without spurious ambiguity, as described in the next section.

\section{Removal of Spurious Ambiguity}
\label{sec:removal}

Let $S$ be a bottom-up shift-reduce transition system that is monotonic.  We show how we can systematically obtain a new transition system $S'$ without spurious ambiguity that is equivalent to $S$, that is, $S'$ parses the same set of trees as $S$.
In essence, this is the main result of this paper, which can be formally stated as follows:

\begin{theorem}
Any transition system $S$ which is bottom-up shift-reduce and monotonic, can always be converted into an {\em equivalent} transition system $S'$ that does not have
spurious ambiguity, such that:

\begin{enumerate}

\item[(i)] 
for each complete computation $\gamma'$ of $S'$ on $w$ there is a complete computation $\gamma$ of $S$ such that $\treeof{\gamma}=\treeof{\gamma'}$; and

\item[(ii)] 
for each complete computation $\gamma$ of $S$ on $w$ there is a complete computation $\gamma'$ of $S'$ such that $\treeof{\gamma}=\treeof{\gamma'}$.

\end{enumerate}
\label{thm1}
\end{theorem}

Next, we describe how $S'$ is created, and give full formal proofs of this theorem in \S\ref{sec:formal}.

\subsection{Stack Symbols}
\label{ssec:stack}

Recall that in $S$ each stack symbol is an integer $i$ representing the word occurrence $a_i$ in the input string.   Each stack symbol in $S'$ is obtained by annotating $i$ with the following Boolean features:

\begin{itemize}

\item 
a feature $i.\fstop$ indicating whether, in the current analysis, the word $a_i$ has collected all of its dependents ($\mytrue$) or it is still seeking some of them ($\myfalse$);

\item 
for each $k$ with $1 \le k \le \degree{S}$, a feature $i.\fredl{k}$ indicating that a left reduction is allowed ($\mytrue$) or forbidden ($\myfalse$) between symbol $i$ and the symbol $k$ positions below $i$ in the stack; 

\item 
for each $k$ with $1 \le k \le \degree{S}$, a feature $i.\fredr{k}$ indicating that a right reduction is allowed ($\mytrue$) or forbidden ($\myfalse$) between symbol $i$ and the symbol $k$ positions below $i$ in the stack.  

\end{itemize}

We now introduce some predicates that will be used later to define the new transition system $S'$. Let $i$ and $j$ be stack symbols of $S'$.  The predicate
$\pbu{i}{j} \equiv \neg i.\fstop \wedge j.\fstop$
indicates whether a bottom-up link from node $i$ to node $j$ is admissible in the current configuration, i.e., whether node $i$ can accept a dependent and node $j$ has already collected all of its dependents. 
Assume that $i$ and $j$ are located at stack positions $p$ and $q$, respectively, with $p > q$.  Then the predicates%
\footnote{Here we are overloading symbols $\fredl{}$ and $\fredr{}$, with related meanings: it will always be clear from the context whether these symbols refer to features or else to predicates.}
\begin{align}
   \pgenredl{i}{j}{p}{q} &\equiv 
      j.\fredl{(p-q)} \wedge \pbu{j}{i} \wedge \reduceleft{p}{q} \in T, \\
   \pgenredr{i}{j}{p}{q} &\equiv 
      j.\fredr{(p-q)} \wedge \pbu{i}{j} \wedge \reduceright{p}{q} \in T 
\end{align}
indicate that reductions $\reduceleft{p}{q}$ and $\reduceright{p}{q}$, respectively, are \termdef{available} in the current configuration, i.e., these reductions can be performed by the parser.  As we will see later, the notion of available reduction plays a crucial role in the construction of $S'$.

\subsection{Transitions}

The basic idea underlying the construction of $S'$ is to perform reductions as early as they become available in a computation, 
according to the notion of available reduction that we have just introduced.
This is implemented as follows.  

We define a \termdef{priority} relation among transitions in $T$ such that, 
in choosing between several reductions that are compatible with some dependency tree, 
we give highest priority to the reduction with its dependent closest to the top of 
the stack. This reduction is necessarily unique, given that in a dependency tree 
each dependent has a unique head.  The shift transitions are always assigned 
the lowest priority.

Note that the priority relation can be seen as a partial order between reductions, but the set
of reductions that are compatible with a given tree is totally ordered, due to the restriction
that a node cannot have more than one head.

In the new transition system $S'$ we simulate $S$ as follows.  
Given a configuration $c'_1$ of $S'$ representing a configuration $c_1$ of $S$, we consider the set $T_{c_1}$ of all transitions from $S$ that are available at $c_1$.  We nondeterministically choose a transition $t \in T_{c_1}$ and simulate it on $c'_1$ under $S'$, moving into a new configuration $c'_2$.  Most important, in $c'_2$ we set the feature of the stack symbols in such a way that all transitions in $T_{c_1}$ that had higher priority than $t$ are now blocked, meaning that no computation spanning from $c'_2$ will ever be able to apply such transitions.   We can now specify our construction.

For a stack symbol $i$ of $S$, we write $i[\mytrue]$ to denote the stack symbol of $S'$ such that $i.\varphi = \mytrue$ for every feature $\varphi$.
For a feature $\varphi$ and a value $v$, we write $j = i[\varphi \leftarrow v]$ if $j.\varphi = v$ and $j.\varphi' = i.\varphi'$ for every other feature $\varphi'$. 
We generalize this notation to a set of features ${\cal F}$, and write $j = i[\varphi \leftarrow v \sep \varphi \in {\cal F}]$ if $j.\varphi = v$ for each $\varphi \in {\cal F}$ and $j.\varphi = i.\varphi$ for each $\varphi \not \in {\cal F}$.
Finally, as a shorthand, we write $i[\varphi \leftarrow v \sep \varphi \in {\cal F}; \varphi' \leftarrow v' \sep \varphi' \in {\cal F}']$ in place of $(i[\varphi \leftarrow v \sep \varphi \in {\cal F}])[\varphi' \leftarrow v' \sep \varphi' \in {\cal F}']$.

The system $S'$ obtained by removing spurious ambiguity from $S$ has
a set of transitions $T'$ including all and only the transitions reported
below, where $\delta$ is $\depth{S}$:
\begin{align}
\conftrans{\transname{\tshiftleaf}^{s}}{(\sigma|i_{\delta}|i_{\delta-1}|\ldots|i_1
, i|\beta, A)}{(\sigma|i_{\delta}'|i_{\delta-1}'|\ldots|i_1'|i' , \beta, A)}
\end{align}
where we let $i' = i[\mytrue]$, and for every $u$ with $1 \leq u \leq \delta$ we
let
\begin{align}
i_u' = 
	i_u[& \fredl{k} \leftarrow \myfalse \sep \pgenredl{i_{u+k}}{i_u}{u+k}{u}; \\
	& \fredr{k} \leftarrow \myfalse \sep \pgenredr{i_{u+k}}{i_u}{u+k}{u}].
\end{align}
Transition $\transname{\tshiftleaf}^{s}$ simulates a shift of $S$.  The superscript $s$ means that the new symbol $i'$ added to the stack has the feature $\fstop$ set to $\mytrue$, that is, we (nondeterministically) guess that $i'$ is now ready for bottom-up reduction.  Since the shift transition has always the lowest priority in $S$, $\transname{\tshiftleaf}^{s}$ blocks any reduction that was available in the antecedent configuration, by setting the features of each $i_q'$, as indicated above. 

We also add to $T'$ a transition $\transname{\tshiftnonleaf}^{\overline{s}}$
defined exactly as $\transname{\tshiftleaf}^{s}$ but with the only difference that we let $i' = (i[\mytrue])[\fstop \leftarrow \myfalse]$, that is, we guess that node $i'$ is still seeking dependents in the current analysis.  

For each \reduceright{p}{q} in $T$, we add to $T'$
\begin{align}
\conftrans{\reducerightleaf{p}{q}}{(\sigma|i_p|i_{p-1}|\ldots|i_1 , \beta, A)}{(\sigma|i'_{p}|\ldots|i'_{q+1}|i'_{q-1}|\ldots|i'_1, \beta, A \cup \{i_p \rightarrow i_q\})}
\end{align}
which can only be applied under the precondition $\pgenredr{i_p}{i_{q}}{p}{q}$.
Here we let $i_p' = i_p[\fstop \leftarrow \mytrue]$, and for every $u$ with $1 \leq u \leq d$ we let
\begin{align}
i_u' = 
    i_u[& \fredl{k} \leftarrow \myfalse \sep u+k < q 
    	\wedge \pgenredl{i_{u+k}}{i_u}{u+k}{u}; \\
	& \fredr{k} \leftarrow \myfalse \sep u < q 
		\wedge \pgenredr{i_{u+k}}{i_u}{u+k}{u}].
\end{align}

As for the shift transition, we also add to $T'$ a transition $\reducerightnonleaf{p}{q}$
defined exactly as ${\reducerightleaf{p}{q}}$ but with $i_p' = i_p[\fstop \leftarrow \myfalse]$.  Reductions $\reducerightleaf{p}{q}$ and $\reducerightnonleaf{p}{q}$ block every reduction $t$ allowable in the antecedent configuration that has priority higher than the reduction $p \rightarrow q$, that is, with a dependent at a position closer to the top of the stack than $q$. 

Similarly to the above, for each \reduceleft{p}{q} in $T$ we add to $T'$
\begin{align}
\conftrans{\reduceleftleaf{p}{q}}{(\sigma|i_p|i_{p-1}|\ldots|i_1 , \beta, A)}{(\sigma|i'_{p-1}|\ldots|i'_1, \beta, A \cup \{i_q \rightarrow i_p\})}
\end{align}
which can only be applied under the precondition $\pgenredl{i_p}{i_{q}}{p}{q}$.
Here we let $i_q' = i_q[\fstop \leftarrow \mytrue]$, and for every $u$ with $1 \leq u \leq d$ we let
\begin{align}
i_u' = 
	i_u[& \fredl{k} \leftarrow \myfalse \sep u+k < p 
		\wedge \pgenredl{i_{u+k}}{i_u}{u+k}{u}; \\
	& \fredr{k} \leftarrow \myfalse \sep u < p \wedge \pgenredr{i_{u+k}}{i_u}{u+k}{u}].
\end{align}

We also add to $T'$ a transition $\reduceleftnonleaf{p}{q}$ defined exactly as ${\reduceleftleaf{p}{q}}$ but with $i_q' = i_q[\fstop \leftarrow \myfalse]$.

The initialization function and final configuration set of $S'$ are like those
of $S$, but we have to specify feature values for the stack symbol
corresponding to the dummy root node $0$: all its features will be $\myfalse$
in the initial configuration, and in final configurations it must have the
$\fredl{k}$ and $\fredr{k}$ features set to $\myfalse$ but $\fstop$ set to
$\mytrue$.



\begin{example}
\label{ex:arcstd2}
If we apply the transformation defined in this section to remove
spurious ambiguity from the arc-standard transition system of Example \ref{ex:arcstd},
we obtain a system $S'$ where the only valid computation for the tree with arcs $\{ {0
\to 2}, {2 \to 1}, {2 \to 3} \}$ is
\[\compbytrans{\init(w)}{\shiftleaf,\shiftnonleaf,\reduceleftnonleaf{2}{1},\shiftleaf,\reducerightleaf{2}{1},\reducerightleaf{2}{1}},\]
\noindent which builds the arcs in the same order as the computation (i) of Example \ref{ex:arcstd}.

It is easy to check that an alternate computation building the arcs in the order of the computation (ii) does not exist in $S'$. Such
a computation would have to start with the transitions $\shiftleaf,\shiftnonleaf,\shiftleaf,\reducerightnonleaf{2}{1}$ 
(the need to use the $s$ or $\overline{s}$ variant of each configuration is uniquely determined by whether nodes have pending
dependents or not).

However, after applying these transitions the parser will be in a configuration $( [0, 1, 2], [] , \emptyset )$ with:

\begin{align}
{0.\fstop=\myfalse}, {0.\fredl{1}=\myfalse}, {0.\fredr{1}=\myfalse}, \\
{1.\fstop=\mytrue}, {1.\fredl{1}=\mytrue}, {1.\fredr{1}=\myfalse}, \\
{2.\fstop=\myfalse}, {2.\fredl{1}=\myfalse}, {2.\fredr{1}=\mytrue}.
\end{align}

\noindent At such configuration, the feature value $2.\fredl{1}=\myfalse$ blocks the left reduction creating the arc $2 \to 1$.
This is so because the $\shiftleaf$ transition that moved the node $3$ to the stack set this value to $\myfalse$,
blocking this left reduction since it could have been executed at that point with higher priority than $\shiftleaf$.  
\end{example}

\section{Formal Properties and Proofs}
\label{sec:formal}

We now proceed to prove that the described transformation for the removal of spurious ambiguity is correct (i.e. prove Theorem~\ref{thm1}).  To do so, we first show that transition systems $S$ and $S'$ defined as in \S\ref{sec:removal} are equivalent, i.e., they assign the same set of trees to any input string.  Afterward, we show that $S'$ has no spurious ambiguity, i.e., different complete computations of $S'$ will always produce different dependency trees.

\subsection{Equivalence of Unambiguous System to Original System}

Let $S$ and $S'$ be defined as in Section \ref{sec:removal}, with associated transition sets $T$ and $T'$, respectively.  To show that $S$ and $S'$ are equivalent, we need to prove that for every input string $w$

\begin{enumerate}

\item[(i)] \label{item:ttos} 
for each complete computation $\gamma'$ of $S'$ on $w$ there is a complete computation $\gamma$ of $S$ such that $\treeof{\gamma}=\treeof{\gamma'}$; and

\item[(ii)] \label{item:stot} 
for each complete computation $\gamma$ of $S$ on $w$ there is a complete computation $\gamma'$ of $S'$ such that $\treeof{\gamma}=\treeof{\gamma'}$.

\end{enumerate}

The proof of~(i) 
is rather straightforward. We show a mapping from the complete computations of $S'$
to the complete computations of $S$ that preserves the associated trees. We define a homomorphism $\myhom$ from $T'$ to $T$ by letting
\begin{align}
\myhom(\reduceleftleaf{p}{q}) = \myhom(\reduceleftnonleaf{p}{q}) = \reduceleft{p}{q},\\
\myhom(\reducerightleaf{p}{q}) = \myhom(\reducerightnonleaf{p}{q}) = \reduceright{p}{q},\\
\myhom({\shiftleaf}) = \myhom({\shiftnonleaf}) = {\shift},\\
\end{align}

and extend it to (complete) computations (recall that we represent a computation by its initial configuration and its sequence of transitions) by letting $\myhom(\compbytrans{c_0}{t_1, \ldots, t_m}) = \compbytrans{c_0}{\myhom(t_1), \ldots, \myhom(t_m)}$.

It is not difficult to see that if $\gamma$ is complete, then $\myhom(\gamma)$ is also complete.  Furthermore, this mapping preserves trees, i.e., for any computation $\gamma$ of $S'$ we have $\treeof{\gamma} = \treeof{\myhom(\gamma)}$, because transitions $t \in T'$ and $\myhom(t) \in T$ create the same arc, if any. This concludes the proof of~(i).  

To prove statement~(ii) above, 
let $\gamma  = c_0, \ldots, c_m = \compbytrans{c_0}{t_1, \ldots, t_m}$ be a complete computation of $S$ for an input string $w$, and let $A_{\gamma}$ be the set of arcs in $\treeof{\gamma}$.  We show that we can always find a computation $\gamma'$ of $S'$ such that $\treeof{\gamma'} = \treeof{\gamma}$. To do this, we introduce below the notion of canonical computations of $S$.  Then we proceed in two steps: first we transform $\gamma$ into a canonical computation $\gamma_f$ of $S$ equivalent to $\gamma$, and then we transform $\gamma_f$ into an equivalent computation $\gamma'$ of $S'$.

Consider a configuration $c_k$, $0 \leq k \leq m$, appearing in $\gamma$.  Let ${\cal R}_{k,\gamma}$ be the set of reductions of $S$ that can be applied to $c_k$, and that are compatible with $D(\gamma)$, i.e., these reductions construct an arc $(h \rightarrow d) \in A_\gamma$.  Here $a_h$ is the head word and $a_d$ is the dependent word, both from $w$.  

Assume that ${\cal R}_{k,\gamma} \neq \emptyset$, and let $t_\rho$ be the reduction in ${\cal R}_{k,\gamma}$ with the highest priority.  This means that $t_\rho$ is the reduction  in ${\cal R}_{k,\gamma}$ with dependent node $d$ placed at the position closest to the top in the stack associated with $c_k$ or, equivalently, the reduction with the largest value of index $d$ in $w$. Note that there cannot be more than one such reduction, due to the single-head constraint in $\treeof{\gamma}$.

We say that $c_k$ is a \termdef{troublesome} configuration in $\gamma$ if $t_{k+1} \neq t_\rho$.  This means that $t_{k+1}$ is either a shift transition, or else a reduction  in ${\cal R}_{k,\gamma}$ that, when applied to $c_k$, creates a dependency link $h' \rightarrow d'$ with $d'<d$, i.e., a reduction with lower priority than $t_\rho$, since node $d'$ will be placed at a deeper position than node $d$ in the stack associated with $c_k$.

We say that a computation of $S$ is in \termdef{canonical} form if it does not contain any troublesome configuration.   This means that, at each configuration $c_k$ of a canonical computation, the reduction in ${\cal R}_{k,\gamma}$ with the highest priority is taken, in case set ${\cal R}_{k,\gamma}$ is not empty.   We now show that for every computation $\gamma$ of $S$ there exists an equivalent  canonical computation $\gamma_f$ of $S$.  We show how to eliminate  the leftmost troublesome configuration in $\gamma$; iteration of this process will always produce a computation where no configurations are troublesome.

Let $c_k$ be the leftmost troublesome configuration in $\gamma$.  We show that we can  build a computation $\gamma_k$ of $S$ which is equivalent to $\gamma$, and such that its first $k$ configurations are not troublesome.  
The transition sequence $t_{k+1}, \ldots , t_m$ can be written in the form
\[
t_{k+1}, t_{k+2}, \ldots, t_{j-1}, t'_{\rho}, t_{j+1}, \ldots, t_m
\]   
where $t'_\rho$ is a reduction creating the same link $h \rightarrow d$ that should have been created by the reduction $t_\rho \in {\cal R}_{k,\gamma}$ with the highest priority. Note that reduction $t'_\rho$ must take place at some $c_j$ in $\gamma$ with $j>k+1$, because $h \rightarrow d$ is in $\treeof{\gamma}$, and this link cannot be present in the arc set associated with $c_k$ (if it were, the reduction $t_\rho$ could not be available at $c_k$ because $d$ would not be in the stack at that configuration).

The sequence $t_{k+1}, \ldots , t_m$ in $\gamma$ can then be replaced (generating the same tree) with
\[
t_\rho, \transftransition{d}{t_{k+1}}, \ldots, \transftransition{d}{t_{j-1}}, t_{j+1}, \ldots, t_m
\]
where $\transftransition{d}{t}$ represents the transition that creates the same arc in a stack where the node $j$ has been removed as $t$ would create in a stack where the node $j$ is present.  Formally, for a transition applied at a configuration $c$ with stack $\sigma | i_p | \ldots | i_q | \ldots | i_1$, we define
$\transftransition{d}{\shift} = \shift$ and
\[
\transftransition{d}{\reduceright{p}{q}} = \begin{cases}
	{\reduceright{p}{q}} \text{ if } i_p > d \text{ and } i_q > d,\\
	{\reduceright{p-1}{q}} \text{ if } i_p < d \text{ and } i_q > d,\\
	{\reduceright{p-1}{q-1}} \text{ if } i_p < d \text{ and } i_q < d.
	\end{cases}
\]

\[
\transftransition{d}{\reduceleft{p}{q}} = \begin{cases}
	{\reduceleft{p}{q}} \text{ if } i_p > d \text{ and } i_q > d,\\
	{\reduceleft{p-1}{q}} \text{ if } i_p < d \text{ and } i_q > d,\\
	{\reduceleft{p-1}{q-1}} \text{ if } i_p < d \text{ and } i_q < d.
	\end{cases}
\]

\noindent
Note that, since $S$ is monotonic, the existence of a transition $t$ implies the existence of $\transftransition{d}{t}$.

The computations $\gamma_k$ and $\gamma$ produce the same tree.
Also, in $\gamma_k$ the first $k$ configurations are not troublesome, since applying the reduction $t_\rho$ at $c_k$ makes $c_k$ not troublesome, and by construction the configurations to the left of $c_k$ in $\gamma_k$ are not troublesome. 

By iteratively applying the above process, we eventually obtain a computation $\gamma_f$ of $S$ such that $\treeof{\gamma_f} = \treeof{\gamma}$. It then remains to show that we can obtain a computation $\gamma'$ of $S'$ with the same associated dependency tree as $\gamma_f$.

Let $\gamma_f = \compbytrans{c_0}{t_1, \ldots, t_m}$ and assume that for each $j$, $1 \leq j \leq m$, transition $t_j$ in $\gamma_f$ applies to configuration $c_{j-1} = (\sigma | i_p | \cdots | i_q | \cdots | i_1, i_0 | \beta, A)$.   The computation $\gamma'$ is obtained as $\gamma' = \compbytrans{c_0}{t'_1, \ldots, t'_m}$, where for each $j$, $t'_j$ is specified as follows.

\begin{itemize}
  \item If $t_j = \reduceright{p}{q}$, then 
  $t'_j$ is $\reducerightnonleaf{p}{q}$ if $A_{\gamma} \setminus (A \cup
  \{(i_p,i_q)\})$ contains a dependency link of the
  form $(i_p,u)$ for some $u$, and $t'_j$ is $\reducerightleaf{p}{q}$ otherwise.
  \item If $t_j = \reduceleft{p}{q}$, then $t'_j$ is
  $\reduceleftnonleaf{p}{q}$ if 
  $A_{\gamma} \setminus (A \cup \{(i_q,i_p)\})$ contains a dependency link of 
  the form $(i_q,u)$ for some $u$, and $t'_j$ is $\reduceleftleaf{p}{q}$ otherwise.
  \item If $t_j = \shift$, then $t'_j$ is $\shiftnonleaf$ if 
  $A_{\gamma} \setminus A$ contains a dependency link of the form $(i_0,u)$ for 
  some $u$, and  $t'_j$ is $\shiftleaf$ otherwise.
\end{itemize}

It is not difficult to see that $\gamma'$ is a valid computation of $S'$ for $w$.  This follows from the fact that the transitions $t'_j$ above satisfy the $\pbu{i}{j}$ predicates in $S'$, and the fact that in $\gamma_f$ reductions are applied in accordance to the priority relation.  We also observe that if $\gamma_f$ is complete then $\gamma'$ is complete as well.  Finally, the fact that $D(\gamma') = D(\gamma_f)$ follows immediately from the above mapping from transitions $t_j$ to transitions $t'_j$.  
This concludes the proof of~(ii)  
and thus the proof of the equivalence of $S$ and $S'$.

\subsection{Non-ambiguity of the Transition System}

To prove that our transformed system $S'$ has no spurious ambiguity, we need to show that different complete computations of $S'$ for $w$ always produce different trees, i.e., if $\gamma_1 \neq \gamma_2$ are complete computations of $S'$ for input string $w$, then $D(\gamma_1) \neq D(\gamma_2)$.

To do so we write $\gamma_1$ as $\alpha c_1 \beta_1$ and $\gamma_2$ as $\alpha c_2 \beta_2$, with $\alpha$ the common
prefix among both computations, and $c_1, c_2$ configurations such that $c_1 \neq c_2$. Note that $\alpha$ cannot be empty, since both
computations must at least have the initial configuration $I(w)$ in common. We call $c_0$ the last configuration in $\alpha$, and $t_1,t_2$ the transitions that produce $c_1,c_2$ (respectively) from $c_0$.  We distinguish four cases below. 

\noindent
Case 1: $t_1$ and $t_2$ are transitions that differ only in the $\fstop$ feature of some new node $u$ in the configuration they produce.  As an example, we have $t_1 = \reduceleftleaf{p}{q}$ and $t_2 = \reduceleftnonleaf{p}{q}$, which differ in the $\fstop$ feature of node $u = q$.   Without loss of generality, we assume $u.\fstop = \mytrue$ in $c_1$, and $u.\fstop = \myfalse$ in $c_2$.  Let $c_0 = (\sigma, \beta, A)$.  Then $D(\gamma_2)$ must contain at least one arc originating from $u$ that is not present in $A$, while $D(\gamma_1)$ cannot contain any arc originating from $u$ that is not already in $A$, because $u.\fstop = \mytrue$ prevents the addition of dependents of $u$ after $t_1$ is executed.  Therefore, $D(\gamma_1) \neq D(\gamma_2)$.

\noindent
Case 2: $t_1$ and $t_2$ are reduce transitions with different head nodes but the same dependent node $u$.  In this $D(\gamma_1) \neq D(\gamma_2)$ follows from the single-head constraint, since the node $u$ will be assigned different heads in $\gamma_1$ and $\gamma_2$, respectively.

\noindent
Case 3: $t_1$ and $t_2$ are reduce transitions involving different dependent nodes.  Suppose that $t_1$ creates the arc $h_1 \rightarrow d_1$ and $t_2$ creates the arc $h_2 \rightarrow d_2$.  Without loss of generality, we assume that $d_1 > d_2$, i.e., $t_1$ has higher priority than $t_2$.  Then $D(\gamma_1)$ contains the arc $h_1 \rightarrow d_1$, but $D(\gamma_2)$ cannot contain this arc, since the system's features block its construction after the application of the transition $t_2$ at configuration $c_0$.

\noindent
Case 4: $t_1$ is a reduce transition and $t_2$ is a shift transition.   The same reasoning of Case~3 applies: the arc $h_1 \rightarrow d_1$ created by $t_1$ cannot appear in $D(\gamma_2)$, because the system's features block its construction after the shift transition $t_2$ is applied.
This concludes the proof that $S'$ does not have spurious ambiguity.

\subsection{Complexity}

Let $S$ be a bottom-up monotonic transition system, and let
$\degree{S} = \delta$.  The construction in \S\ref{sec:removal}
adds $2\delta + 1$ binary features to each stack symbol of $S$.
This results in $2^{2\cdot\delta+1}$ new symbols in $S'$ for each
stack symbol of $S$.   While for projective dependency parsing we 
have $\delta = 1$, degree larger than one is needed 
in non-projective parsing.  However, it has been observed 
by \newcite{attardi2006experiments} that most of the non-projective 
trees in the CoNLL data can be parsed with $\delta = 2$ or $3$.  
This means that, in practical cases, the blow-up of stack symbols 
by our construction can be considered a small constant.

To discuss a concrete application, consider the non-projective system $S$ of \citep{attardi2006experiments}, also shown in Example~\ref{ex:attardi}, restricted 
to $\delta = 2$, which is still heavily affected by spurious 
ambiguity.  We have applied the construction in 
\S\ref{sec:removal} to $S$ with some ad-hoc optimization 
of the features for that system, resulting in a new system $S'$ 
with a blow-up of stack symbols of $2^{\delta+1} = 8$.  This 
means that we can apply to $S'$ the inside/outside algorithm 
presented in \cite{cohen2011exact}, working in 
time $\order{\size{w}^7}$ for an input string $w$, with an extra 
hidden constant of $8$.

\section{Experiments}
\label{sec:experiment}

As mentioned earlier, transition-based dependency parsing uses an oracle to convert training data which consists of pairs of sentences
and dependency trees to pairs of sentences with shift-reduce sequences, in order to sidestep the issue of spurious ambiguity. The new training data is then used to train multi-class
classifiers. In several cases, oracles are based on heuristics and are incomplete.
The oracle that is provided in the DeSR dependency parsing package,%
\footnote{\url{http://desr.sourceforge.net/}.} 
which is based on the parser from
\namecite{attardi2006experiments}, is an example for such incomplete
heuristics.

We compared the coverage of Attardi's oracle, restricted to transitions of
degree at most $2$, to the oracle of an equivalent transition system without
spurious ambiguity.\footnote{Note that the algorithm implemented in the latest version of DeSR,
which we used for these experiments, differs from the description
provided in \namecite{attardi2006experiments} and Example~\ref{ex:attardi} in that
\reduceleft{3}{1} and \reduceright{3}{1} transitions push a node from
the stack back to the buffer after reducing. This does not affect our method to remove
spurious ambiguity, which is correct both for the version described in
\namecite{attardi2006experiments} and for the latest implementation of Attardi's parser.}
Our findings are given in Table~\ref{table:oracle}. As theoretically guaranteed, there were no cases where Attardi's parser recognized
a tree using transitions of degree 2, and our oracle did not recognize it. The reverse, however, holds quite often.

\begin{table}
\begin{center}
\begin{tabular}{l|r|r|r}
Language & Size & Attardi & This paper \\
\hline
Arabic & 1,460 & 27 & 2 \\
Bulgarian & 12,823 & 47 & 36 \\
Czech & 72,703 & 1,334 & 602 \\
Danish & 5,190 & 179 & 159 \\
Dutch & 13,349 & 1,448 & 1,018 \\
German & 39,216 & 2,140 & 1,538 \\
Japanese & 17,044 & 121 & 45 \\
Portuguese & 9,071 & 295 & 203 \\
Slovene & 1,534 & 48 & 27 \\
Spanish & 3,306 & 11 & 10 \\
Swedish & 11,042 & 197 & 105 \\
Turkish & 4,997 & 208 & 102
\end{tabular}
\end{center}
\caption{\label{table:oracle}Coverage of Attardi's oracle versus the coverage of our oracle for various treebanks from the CoNLL 2006 data sets \citep{buchholz2006conll}. ``Size'' denotes
the number of sentences in the treebank (we used the training portion only), ``Attardi'' denotes the number of sentences that Attardi's oracle could not parse and ``this paper'' denotes
the number of parse trees that our oracle could not parse.}
\end{table}

\section{Discussion}
\label{sec:discussion}

We note that monotonic bottom-up shift-reduce transition systems can be made
probabilistic and generative, in a manner similar to \newcite{cohen2011exact}. The issue with spurious ambiguity is especially crucial with generative models in the
unsupervised setting, when using algorithms such as the expectation-maximization (EM) algorithm.
\newcite{cohen2011exact} describe 
an EM algorithm for the 
system from \newcite{attardi2006experiments}, which can 
be extended to any monotonic bottom-up
transition system. The EM algorithm they describe can be further extended to monotonic bottom-up transition systems after removal of spurious ambiguity (as we describe in this
paper), making these systems readily available for transition-based unsupervised learning for dependency parsing.

\section{Conclusion}
\label{sec:conclusion}

We provided a principled treatment to the issue of spurious ambiguity in transition-based dependency parsing. We defined a large class of transition systems, which we call
monotonic bottom-up shift-reduce transition systems, that cover existing systems
such as
the arc-standard parser of \newcite{nivre2008algorithms} and the non-projective 
parser of \newcite{attardi2006experiments}, 
as well as systems in which reductions affect elements at positions in the stack deeper than the topmost element \citep{goldberg-elhadad:2010:NAACLHLT}.
We then showed how to eliminate spurious ambiguity from these systems. Our technique has applications for unsupervised and supervised dependency parsing.
The transition model that we present can be used as a substitute for models such as the dependency model with valence that have long been used for
dependency grammar induction \citep{klein-04b,cohen-10d,spitkovsky-10}.

In this paper we have discovered some sufficient conditions under which spurious ambiguity can be removed from bottom-up dependency transition systems, which we hope are as ``tight'' as possible. However, our technique does not work for all dependency transition systems, and it remains an open problem to show whether removal of spurious ambiguity can be carried out in the general case. There might as well be dependency parsing strategies for which removal of spurious ambiguity is not only difficult, but simply impossible.  A similar scenario is observed, for instance, for structural ambiguity in context-free grammars, where some context-free languages can only be generated using ambiguous context-free grammars; see for instance~\namecite{HO79}.

\bibliography{mcqm,bibliography}

\end{document}